\documentclass{article}

\usepackage{arxiv}

\usepackage[utf8]{inputenc} 
\usepackage[T1]{fontenc}    
\usepackage{hyperref}       
\usepackage{url}            
\usepackage{booktabs}       
\usepackage{amsfonts}       
\usepackage{nicefrac}       
\usepackage{microtype}      
\usepackage{lipsum}
\usepackage{graphicx}
\graphicspath{ {./images/} }

\usepackage{tabularray}

\usepackage[normalem]{ulem}
\usepackage{array}
\usepackage{ragged2e}
\usepackage{multirow}
\usepackage{colortbl}
\usepackage{caption}
\usepackage{subcaption}

\title{Reinforcement Learning of Large Language Models for Interpretable Credit Card Fraud Detection}

\author{
  \textbf{Cooper Lin$^{1,*}$, Yanting Zhang$^{1,*}$, Maohao Ran$^{1,2}$, Wei Xue$^{1}$,} \\
  \textbf{Hongwei Fan$^{3}$, Yibo Xu$^{1}$, Zhenglin Wan$^{4}$, Sirui Han$^{1}$, Yike Guo$^{1}$, Jun Song$^{1,2,\dagger}$} \\
  \\
  $^1$Hong Kong University of Science and Technology, $^2$Hong Kong Baptist University\\ $^3$Imperial College London, $^4$National University of Singapore\\
}
\date{}

\begin{document}
\maketitle
{\let\thefootnote\relax\footnotetext{$^*$Equal contribution.}}
{\let\thefootnote\relax\footnotetext{$^\dagger$Corresponding author: junsong@hkbu.edu.}}
\begin{abstract}
E-commerce platforms and payment solution providers face increasingly sophisticated fraud schemes, ranging from identity theft and account takeovers to complex money laundering operations that exploit the speed and anonymity of digital transactions. However, despite their theoretical promise, the application of Large Language Models (LLMs) to fraud detection in real-world financial contexts remains largely unexploited, and their practical effectiveness in handling domain-specific e-commerce transaction data has yet to be empirically validated. To bridge this gap between conventional machine learning limitations and the untapped potential of LLMs in fraud detection, this paper proposes a novel approach that employs Reinforcement Learning (RL) to post-train lightweight language models specifically for fraud detection tasks using only raw transaction data. We utilize the Group Sequence Policy Optimization (GSPO) algorithm combined with a rule-based reward system to fine-tune language models of various sizes on a real-life transaction dataset provided by a Chinese global payment solution company. Through this reinforcement learning framework, the language models are encouraged to explore diverse trust and risk signals embedded within the textual transaction data, including patterns in customer information, shipping details, product descriptions, and order history. Our experimental results demonstrate the effectiveness of this approach, with post-trained language models achieving substantial F1-score improvements on held-out test data. Our findings demonstrate that the observed performance improvements are primarily attributable to the exploration mechanism inherent in reinforcement learning, which allows models to discover novel fraud indicators beyond those captured by traditional engineered features.
\end{abstract}


\section{Introduction}

The rapid expansion of digital commerce has transformed global economic landscapes, enabling unprecedented transaction volumes and cross-border payment flows. However, this growth has been accompanied by a parallel surge in fraudulent activities, with online payment fraud causing billions of dollars in losses annually to merchants, financial institutions, and consumers worldwide \cite{dastidar2024machine}. E-commerce platforms and payment solution providers face increasingly sophisticated fraud schemes, ranging from identity theft and account takeovers to complex money laundering operations that exploit the speed and anonymity of digital transactions. Traditional rule-based fraud detection systems, which rely on predefined thresholds and static rules, have proven inadequate against adaptive fraudsters who continuously evolve their tactics to circumvent detection mechanisms. This arms race between fraudsters and fraud prevention systems has intensified the demand for more intelligent, adaptive, and robust detection methodologies capable of identifying subtle patterns and anomalies in vast transaction datasets.

\begin{figure}[t]
    \centering
    \includegraphics[width=\linewidth]{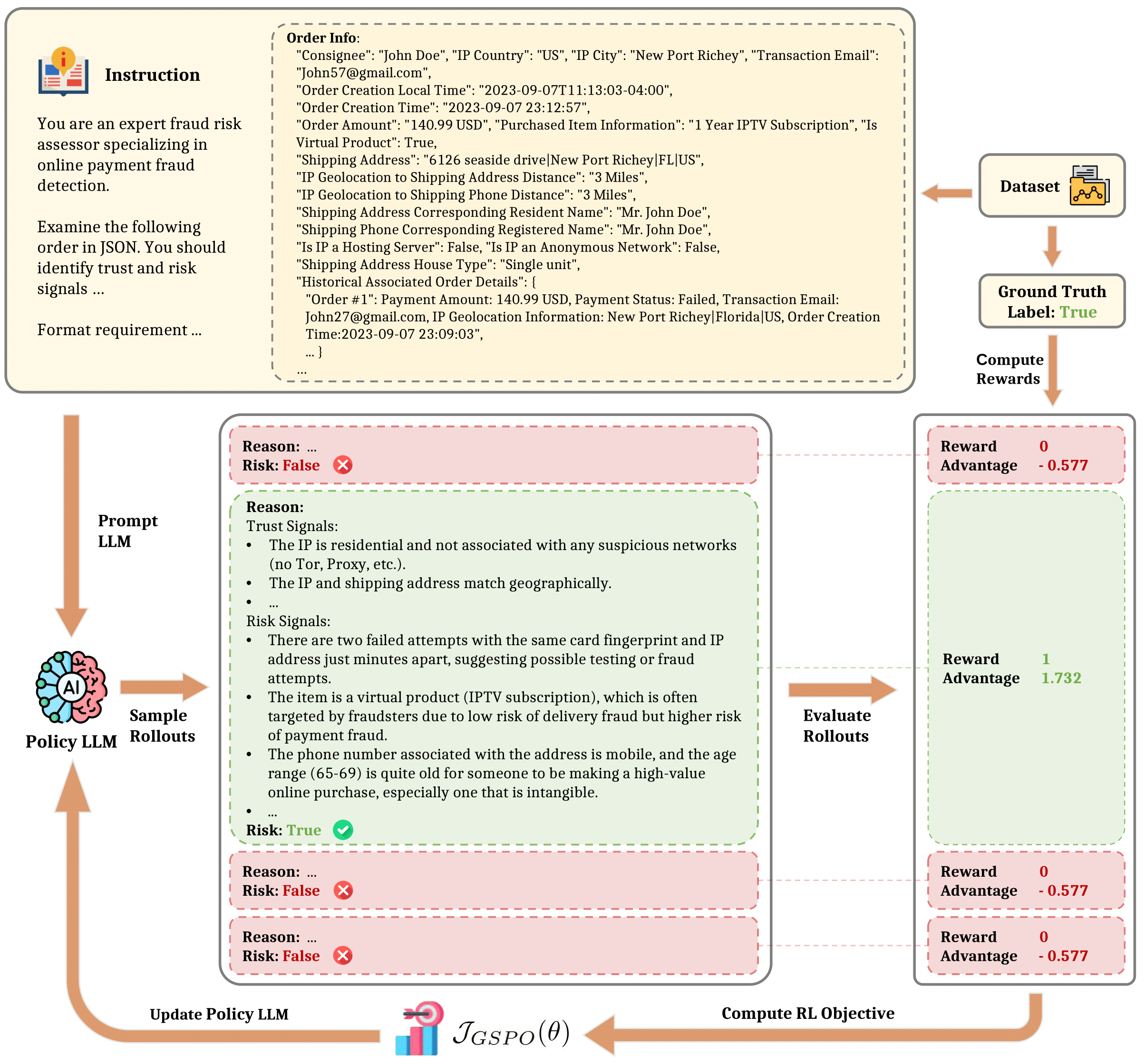}
    \caption{\textbf{Overview of the GSPO reinforcement learning framework for fraud detection.} For each input transaction, the policy model samples a group of candidate responses (rollouts). These outputs are evaluated against the ground truth label to compute rewards and derive group-relative advantage estimates. Finally, these advantages drive the GSPO optimization objective, updating the model parameters to reinforce accurate reasoning and verdicts. Note that the "Order Info" panel displays a representative subset of the complete feature set for brevity.}
    \label{fig:framework}
\end{figure}

Machine learning has emerged as the dominant paradigm for automated fraud detection, with gradient boosting algorithms such as Neural Networks \cite{alghofaili2020financial, chen2021deep, udayakumar2023deep}, XGBoost \cite{hajek2023fraud, noviandy2023credit, baisholan2025fraudx}, and LightGBM \cite{zhao2024improved, xiao2025application} achieving remarkable success in production environments due to their ability to handle tabular data efficiently and deliver high predictive accuracy. These models excel at learning complex non-linear relationships from structured features such as transaction amounts, timestamps, device fingerprints, and historical behavioral statistics. Despite their effectiveness, conventional machine learning approaches face fundamental limitations that constrain their fraud detection capabilities. First, they \textbf{require extensive feature engineering efforts} to transform raw transaction data into meaningful numerical representations, a process that demands domain expertise and may overlook critical patterns not explicitly encoded. Second, these models are \textbf{inherently designed for structured, tabular data and cannot effectively process or extract insights from unstructured textual information}—such as customer names, email addresses, shipping addresses, product descriptions, and merchant details—which often contain rich behavioral and social signals indicative of fraudulent intent. The inability to comprehend semantic relationships, linguistic patterns, and contextual nuances in textual data represents a significant blind spot in current fraud detection systems, leaving valuable information unexploited and potentially allowing sophisticated fraud schemes to evade detection. Third, and most critically, \textbf{existing machine learning approaches fundamentally lack interpretability}—a deficiency attributable either to the opacity of the underlying model architecture (e.g., deep neural networks) or to the aggregated nature of engineered features that obscures the relationship between raw inputs and model predictions.

Recent advances in Large Language Models (LLMs) present a promising avenue to address these critical gaps in fraud detection capabilities. Pre-trained on vast corpora of text data, LLMs such as GPT \cite{gpt5}, Claude \cite{claude4_5}, and Qwen \cite{yang2025qwen3} have demonstrated remarkable abilities to understand complex linguistic patterns, semantic relationships, and contextual information across diverse domains. Unlike traditional machine learning models that treat textual fields as categorical variables or require manual encoding, \textbf{LLMs can natively process and comprehend unstructured text, potentially uncovering behavioral patterns and social signals} embedded in customer names, contact information, shipping addresses, and transaction descriptions \cite{thapa2025large, ferrag2025llm}. For instance, LLMs may detect suspicious patterns such as anomalous email address formats, inconsistencies between shipping and billing information, unusual product purchase combinations, or subtle linguistic cues that indicate coordinated fraud rings operating across multiple accounts. Beyond their textual comprehension capabilities, \textbf{LLMs offer a fundamental advantage in interpretability}: their reasoning process unfolds in natural language, enabling human analysts to directly examine the rationale behind fraud predictions rather than reverse-engineering opaque numerical features or model weights. \textbf{This transparency is particularly valuable in financial contexts} where regulatory compliance, audit requirements, and the need to explain decisions to stakeholders demand clear justification for flagging transactions as fraudulent. Furthermore, the inherent ability of LLMs to perform in-context learning and adapt to new fraud patterns with minimal additional fine-tuning data positions them as potentially more flexible and resilient solutions compared to conventional models that require retraining on extensive labeled datasets. However, despite their theoretical promise, the application of LLMs to fraud detection in real-world financial contexts remains largely unexplored, and their practical effectiveness in handling domain-specific e-commerce transaction data has yet to be empirically validated.

To bridge this gap between conventional machine learning limitations and the untapped potential of LLMs in fraud detection, this paper proposes a novel approach that \textbf{employs Reinforcement Learning (RL) to post-train open-source language models specifically for fraud detection tasks}. We utilize the Group Sequence Policy Optimization (GSPO) algorithm \cite{zheng2025group} combined with a rule-based reward system to fine-tune language models of various sizes on a real-world e-commerce transaction dataset provided by a Chinese global payment solution company. A key advantage of our methodology is its \textbf{minimal annotation requirement}: we only need the raw transaction information as model input and binary fraud labels for reward calculation, while the reasoning process and fraud predictions are generated autonomously by the policy models during RL training. Through this reinforcement learning framework, the language models are encouraged to \textbf{explore diverse trust and risk signals embedded within the textual transaction data}, including patterns in customer information, shipping details, product descriptions, order history, and their interrelationships. The reward-driven optimization process inherently identifies which signals are genuinely predictive of fraudulent behavior and updates model parameters accordingly in an end-to-end fashion, enabling the discovery of non-obvious fraud patterns and facilitating adaptation to evolving fraud tactics without explicit feature engineering.

Our experimental results demonstrate the effectiveness of this approach, with post-trained language models achieving \textbf{substantial F1-score improvements on held-out test data} using only raw textual transaction information as input. Notably, the models attain these performance gains while maintaining computational efficiency: during RL training, they learn to reduce their average generation length, producing more concise predictions without sacrificing accuracy. This efficiency is particularly crucial for fraud detection systems that must operate in real-time environments where millisecond-level latency can impact user experience and transaction throughput. Through comprehensive ablation studies, we further reveal that the observed performance improvements are primarily attributable to the exploration mechanism inherent in reinforcement learning, which allows models to discover novel fraud indicators beyond those captured by traditional engineered features. Collectively, our methodology and empirical findings directly address the identified research gap, demonstrating that \textbf{appropriately trained language models can effectively leverage unstructured textual data for fraud detection} and highlighting the significant potential for LLM-based fraud detection systems in real-world financial applications.

\section{Preliminaries}

\subsection{LLM Alignment}

After pre-training on large text corpora, Large Language Models (LLMs) require further refinement to align their behavior with specific tasks such as credit card fraud detection. Two primary alignment methods are Supervised Fine-Tuning (SFT) and Reinforcement Learning (RL) \cite{wang2024comprehensive}.

Supervised Fine-Tuning trains models using curated datasets of input-output pairs, minimizing the difference between model predictions and labeled correct answers. SFT is effective when a large, high-quality labeled dataset is available, and the task has well-defined, objective answers. However, in the context of credit card fraud detection through verbal reasoning, creating such a dataset presents significant challenges. First, creating a training corpus requires fraud experts to manually analyze thousands of transactions and provide detailed explanations—a process that is prohibitively expensive and time-intensive. Second, expert annotations may be inconsistent or reflect known fraud patterns rather than capturing emerging tactics. Third, SFT models tend to memorize training examples, potentially limiting their ability to generalize to novel fraud schemes. This approach risks training a model that excels at identifying known patterns but is less adept at discovering new tactics and is also biased towards certain unverified, potentially misleading signals.

Reinforcement Learning offers an alternative approach better suited to our task. Rather than learning from fixed correct answers, the policy LM samples a series of reasoning for each transaction and receives reward signals indicating detection accuracy. This approach offers four key advantages. First, it eliminates the need for extensive expert-annotated reasoning datasets, as the response used for updating the policy model is generated by itself in online rollouts. Second, it enables exploration-exploitation trade-offs: the model is incentivized to explore diverse reasoning strategies while exploiting successful ones, potentially discovering novel fraud indicators beyond human-curated patterns. Third, because rewards are based solely on detection accuracy, the model learns patterns directly from outcomes rather than relying on potentially biased expert interpretations. Fourth, RL has demonstrated superior generalization to out-of-distribution data, which is a critical property given that fraudsters continuously adapt their tactics. For these reasons, we employ RL-based alignment for our fraud detection model.

\subsection{Reinforcement Learning with Group-Average Baseline}\label{chap:pre_rl}

In reinforcement learning for LLM alignment, the model generates outputs and receives rewards based on task performance. A critical challenge in this process is the high variance in reward signals, which can destabilize training and slow convergence. To address this, RL algorithms typically employ a baseline—a reference value subtracted from the raw reward to produce a more stable learning signal \cite{weaver2013optimal}. Consider a scenario where the model generates reasoning for multiple transactions, all receiving positive rewards of varying magnitudes. Without a baseline, the model would reinforce all these outputs proportionally to their absolute rewards. However, what matters for learning is not the absolute reward value, but rather which outputs performed better or worse \textit{relative to expectations}. By subtracting a baseline, we obtain advantage values that indicate whether an output exceeded or fell short of typical performance, providing clearer guidance for model updates.

Formally, given state $s$ and action $a$, instead of using raw rewards $r(s, a)$ to update the model, we use advantages $A(s, a) = r(s, a) - b(s)$, where b is the baseline, a scaler function dependent only on $s$. This centered signal reduces variance while maintaining the correct gradient direction, as the baseline affects all actions equally and does not bias the policy gradient. A common baseline approach uses a separate value network—a learned model that estimates expected rewards for given inputs, e.g., Proximal Policy Optimization (PPO) \cite{schulman2017proximal}. While effective, this approach adds computational overhead and requires training an additional model alongside the policy. For LLM alignment tasks, a simpler and more efficient alternative has gained prominence: the group-average baseline.

A notable algorithm that implements this approach is Group Relative Policy Optimization (GRPO) \cite{shao2024deepseekmath}, which generates $G$ output samples $\{y_i\}_i^G$ for each input $x$ and uses their normalized reward within the group as the baseline. Specifically, given a query set $\mathcal{D}$, GRPO optimizes the following objective:

\begin{equation}
    \mathcal{J}_{GRPO}=\mathbb{E}_{x\sim\mathcal{D},\{y_i\}_{i=1}^G\sim\pi_{\theta old}} \left[ \frac{1}{G}\sum_{i=1}^G\frac{1}{|y_i|}\sum_{t=1}^{|y_i|}\mathrm{min}\left(w_{i, t}(\theta)A_{i,t}, \mathrm{clip}\left(w_{i, t}(\theta), 1-\epsilon, 1+\epsilon\right)A_{i,t}\right)\right]
\end{equation}

where the importance ratio $w_{i,t}^\theta$ and the group-averaged advantage estimation $A_{i,t}$ of token $y_{i,t}$ are:

\begin{equation}
    w_{i,t}(\theta)=\frac{\pi_\theta(y_{i,t}|x,y_{i,<t})}{\pi_{\theta old}(y_{i,t}|x,y_{i,<t})}, \quad
    A_{i,t}=\frac{r(x, y_i)-\mathrm{mean}\left(\{r(x,y_i)\}_{i=1}^G\right)}{\mathrm{std}\left(\{r(x,y_i)\}_{i=1}^G\right)}
\end{equation}

Despite GRPO's success on various RL tasks, two characteristics make it suboptimal for real-time fraud detection \cite{liu2025understanding}. First, the length normalization term $|y_i|$ in GRPO's objective inherently encourages longer outputs during training. As the model learns, generated explanations tend to grow in length, increasing inference latency, which is a critical concern for fraud detection systems that must process transactions in real-time. Second, GRPO exhibits a structural inconsistency: importance weights are computed at the token level, while rewards reflect the quality of entire output sequences. This granularity mismatch complicates credit assignment in our setting, where detection accuracy depends on the complete reasoning chain rather than individual tokens, potentially impeding effective gradient propagation.

\section{Methodology}\label{chap:RL}

\begin{figure}[t]
    \centering
    \includegraphics[width=0.95\linewidth]{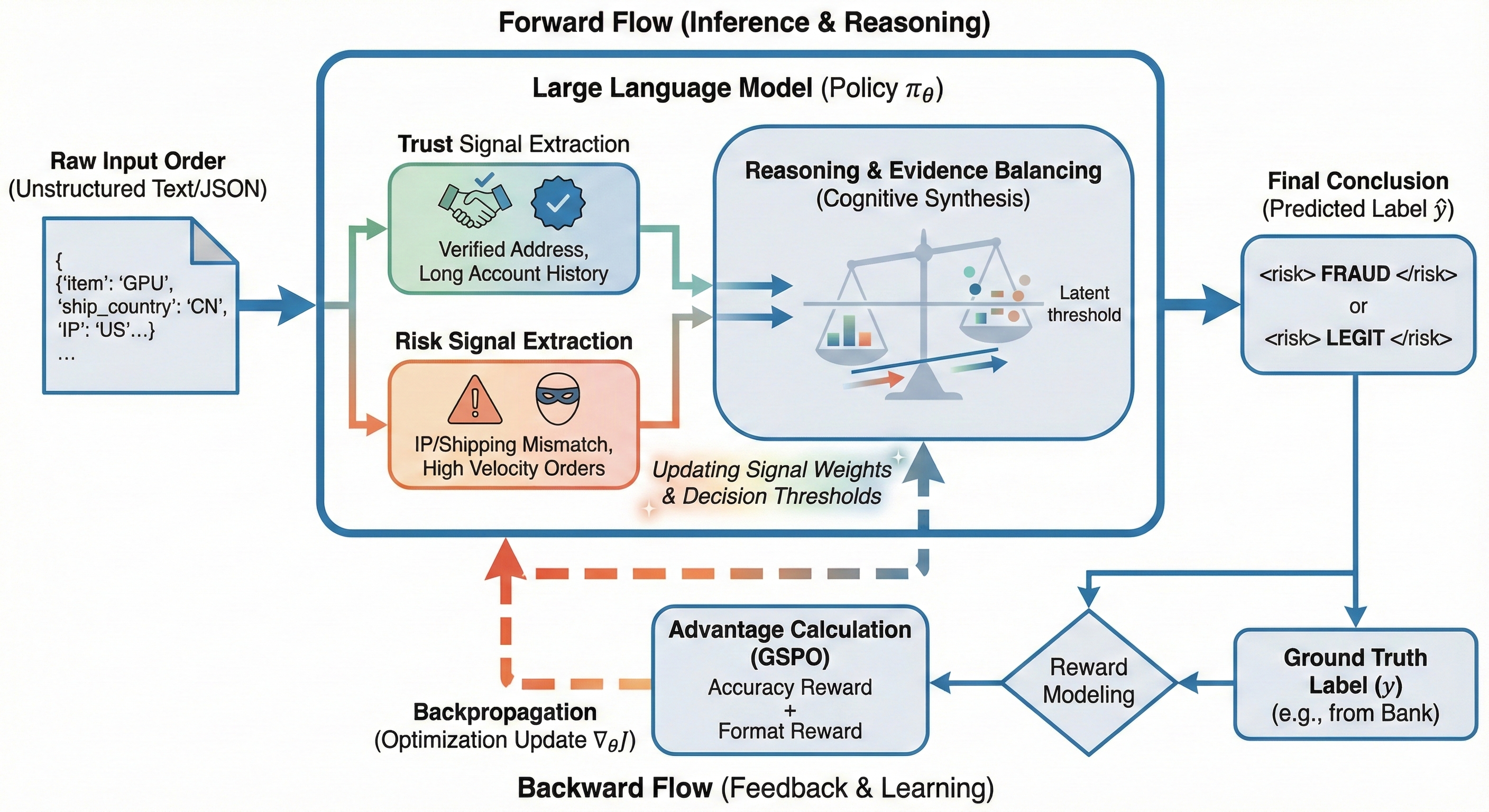}
    \caption{\textbf{Schematic of the reinforcement learning cycle for fraud detection.} The \textbf{forward path} illustrates the model's reasoning chain: extracting diverse trust and risk signals from raw inputs and synthesizing evidence against a latent threshold to render a verdict. The \textbf{backward path} demonstrates optimization: reward signals derived from ground truth labels drive backpropagation, updating parameters to implicitly learn signal relevance and calibrate domain-specific decision boundaries.}
    \label{fig:learning}
\end{figure}

Conventional machine learning approaches to credit card fraud detection face two fundamental limitations: they rely exclusively on numeric features while overlooking textual and behavioral patterns, and they suffer from limited interpretability due to complex feature engineering aggregations. To address these challenges, we employ Large Language Models (LLMs) to perform classification through natural language reasoning. However, directly applying general-purpose LLMs to fraud detection presents significant obstacles. These models lack domain-specific knowledge about which trust or risk signals constitute meaningful fraud indicators and how sensitively such signals should influence classification decisions. Moreover, the relevance and sensitivity of these signals vary across different risk control scenarios. To align LLMs with fraud detection requirements while maintaining low response latency through concise outputs, we employ reinforcement learning as a post-training technique to enhance their task-specific performance.

\subsection{Reinforcement Learning Algorithm}

\begin{figure}[t]
    \centering
    \includegraphics[width=\linewidth]{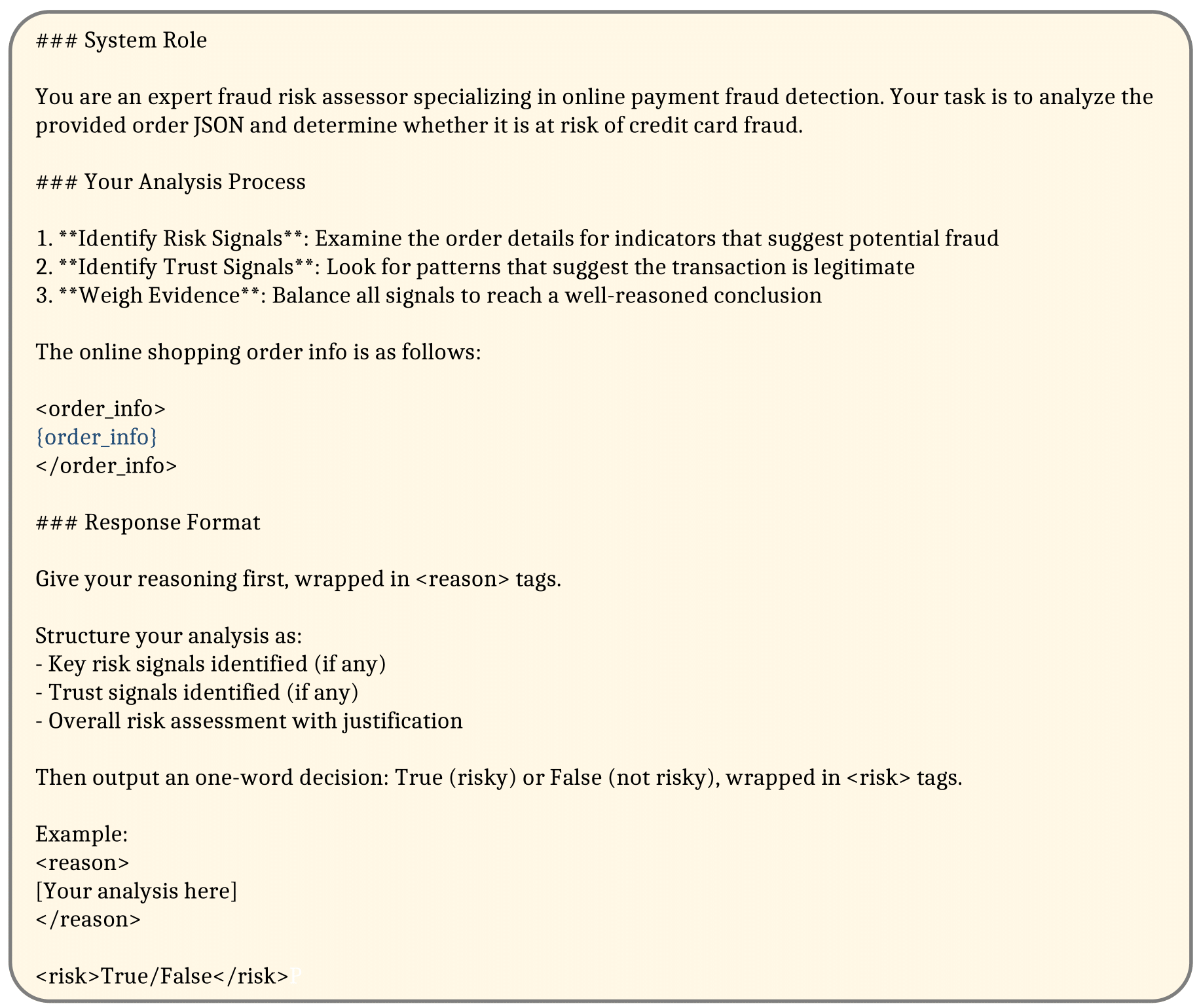}
    \caption{Prompt template for our fraud detection LMs.}
    \label{fig:prompt}
\end{figure}

As discussed in Section \ref{chap:pre_rl}, reinforcement learning approaches utilizing group-relative baselines, such as the Group Relative Policy Optimization (GRPO) algorithm \cite{shao2024deepseekmath}, have demonstrated promising results in alignment tasks. However, as mentioned, GRPO's token-level objective formulation and completion length normalization scheme introduce potential complications in our application, potentially leading to suboptimal performance. Most notably, the normalization scheme tends to incentivize increasingly longer completions, which poses significant latency concerns for fraud detection systems that require real-time responses. To mitigate these issues, \textbf{we adopt the Group Sequence Policy Optimization (GSPO) algorithm} \cite{zheng2025group}, which employs a \textit{sequence-level} importance weighting mechanism in its objective function, naturally aligning with sequence-level reward structures in our fraud detection task. The GSPO optimization objective is formulated as follows:

\begin{equation}
    \mathcal{J}_{GSPO}(\theta)=\mathbb{E}_{x\sim\mathcal{D},\{y_i\}_{i=1}^G\sim\pi_{\theta old}}\left[\frac{1}{G}\sum_{i=1}^G\mathrm{min}\left(s_i(\theta) A_i, \mathrm{clip}(s_i(\theta),1-\epsilon, 1+\epsilon)A_i\right)\right]
\end{equation}

where the sequence-level importance weight is defined as:

\begin{equation}
    s_i(\theta)=\mathrm{exp}\left(\frac{1}{|y_i|}\sum_{i=1}^{|y_i|}\mathrm{log}\frac{\pi_\theta(y_{i,t}|x,y_{i,<t})}{\pi_{\theta old}(y_{i,t}|x,y_{i,<t})}\right)
\end{equation}

and the group-relative advantage estimation $A_i$ is the same as GRPO. Note that the length normalizer $|y_i|$ has been removed from the outer objective, which prevents the model from diluting negative advantages by generating increasingly longer completions. However, this normalizer is retained within the sequence importance weight $s_i(\theta)$ to reduce variance and maintain $s_i(\theta)$ within a consistent numerical range, thereby preventing dramatic fluctuations that could arise from short completions.

In our fraud detection setup, the input prompt $x$ consists of three components: the fraud detection task instruction, the e-commerce order information, and the output format requirements. The sampled output $y_i$ comprises the model's reasoning (i.e., the identified trust and risk signals) followed by its final fraud verdict.

\subsection{Reward Modeling}

\begin{figure}[t]
    \centering
    \includegraphics[width=\linewidth]{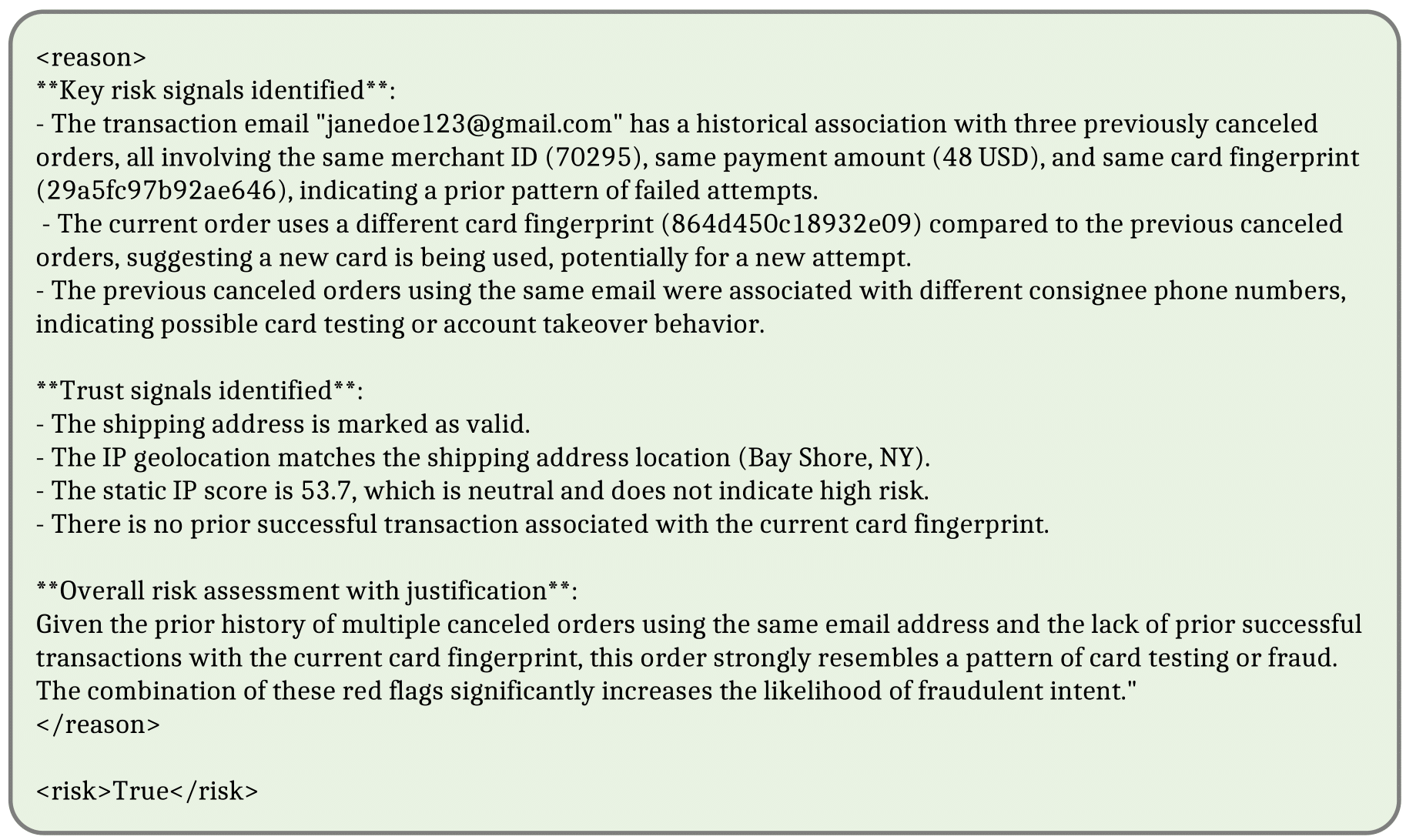}
    \caption{An example output generated by the trained Qwen3-14B on the test set.}
    \label{fig:example}
\end{figure}

The reward function provides training signals that fundamentally determine the optimization direction of reinforcement learning. Since fraud detection is highly result-sensitive, with performance almost entirely determined by the final verdict, \textbf{we implement a rule-based outcome reward system comprising two components}:

\begin{itemize}
    \item \textbf{Accuracy Reward}: This component evaluates whether the model's final fraud risk verdict aligns with the ground truth label provided by the bank. The verdict must be deterministic and binary (fraudulent or legitimate). To enable reliable rule-based extraction and verification, the model must output its final answer in a specific format (e.g., wrapped in \texttt{<risk>} tags).

    \item \textbf{Format Reward}: Beyond accuracy, this component incentivizes the model to first identify a series of key trust and risk signals from the input order that inform its decision. These signals are instructed to be wrapped in \texttt{<reason>} tags, ensuring the model articulates its reasoning before rendering a verdict.
\end{itemize}

In our experiment, we weight the accuracy reward 2.5 times higher than the format reward, because we observe that during training, the model consistently achieves the format reward, making it a less informative training signal. By scaling up the accuracy reward, we prioritize optimization of the model's fraud detection performance, which is the primary objective. We deliberately avoid process rewards since we only have access to reliable ground truth fraud labels, making it extremely difficult to implement proper, verifiable intermediate checkpoints. Meanwhile, empirical studies show that outcome rewards generally yield favorable results in LLM reinforcement learning, whereas process rewards may encourage reward hacking, as partial credits can be exploited throughout the reasoning chain \cite{gao2024designing, guo2025deepseek}.

\subsection{Prompt and Completion Format}

In our fraud detection framework, we instruct LLMs to provide structured reasoning before rendering a final verdict on whether an order poses a credit card fraud risk. Specifically, the model must first identify two types of evidence from the input transaction data:

\begin{itemize}
    \item \textbf{Risk signals}: Features or patterns that appear suspicious and suggest potential fraud (e.g., geographic inconsistencies, unusual purchasing behavior, mismatched identity information).
    \item \textbf{Trust signals}: Features or patterns that appear legitimate and support the order's authenticity (e.g., verified address correspondence, established customer history, consistent payment patterns).
\end{itemize}

After identifying these signals, the model must synthesize all evidence to reach a balanced, justified conclusion. Our prompt design is deliberately open-ended: we do not prescribe what constitutes valid trust or risk signals, nor do we provide an exhaustive enumeration of potential fraud indicators. This design choice serves a strategic purpose in reinforcement learning. By leaving the signal identification unconstrained, we encourage the model to explore diverse features it deems relevant during training. Through the reward system and gradient-based optimization, the model can discover and retain truly informative signals in an end-to-end manner, rather than being limited to predefined heuristics. This approach allows the model to potentially uncover non-obvious fraud patterns and adapt to evolving fraud tactics that might not be captured in hand-crafted rules. The complete prompt template and an example output are shown in Figure \ref{fig:prompt} and \ref{fig:example}, respectively.

\section{Experiments}

\subsection{Dataset}

We conducted experiments on an e-commerce transaction dataset provided by a Chinese global payment solution company, spanning transactions from June 2023 to June 2024. Each transaction record is in a JSON format and contains comprehensive information across multiple categories:

\begin{itemize}
    \item \textbf{Transaction Details}: Basic order information, including consignee name, order amount, creation timestamp (both local and UTC), purchased item details (product name, SKU, quantity, price), and whether the product is virtual.
    \item \textbf{Geographic and Network Information}: Transaction IP address with geolocation data (country, city, state), IP characteristics (hosting server, anonymous network, VPN, proxy, Tor status), IP address type (e.g., cellular, residential), IP registration country, device information, and static IP score. These features are mainly acquired through the MaxMind APIs.
    \item \textbf{Payment Information}: Card fingerprint, card brand (e.g., Visa, Mastercard), issuing bank, card type (credit/debit), and card characteristics (business, prepaid, or virtual card status).
    \item \textbf{Shipping Information}: Shipping address with detailed components (street, city, state, country), consignee phone number with country code and line type (mobile/landline), shipping address type (residential/business, freight forwarder status), and address validity verification.
    \item \textbf{Identity Verification Features}: Transaction email address, days since first observation of the email, correspondence between shipping address/phone/email and registered names, phone number validation, and registrant age range (when available). These features are mainly acquired through the Ekata APIs.
    \item \textbf{Distance and Consistency Check}: Geographic distance metrics (IP geolocation to shipping address distance, IP geolocation to phone number distance), consistency checks (shipping phone matches address, email validity). These features are mainly acquired through the Ekata APIs.
    \item \textbf{Historical Context}: Historical associated orders linked by shared identifiers (email, phone number, card fingerprint, IP address), including their payment amounts, statuses (successful, canceled, failed), and creation timestamps.
\end{itemize}

\begin{table}[h]
\centering
\caption{Data Overview}
\label{tab:data}
\begin{tblr}{
  column{even} = {r},
  column{3} = {r},
  column{5} = {r},
  hline{1,4} = {-}{0.08em},
  hline{2} = {-}{0.05em},
}
             & \# of Samples & Legit. Orders & Fraud. Orders & Time Range       \\
Training Set & 4900          & 2586 (51.8\%) & 2314 (47.2\%) & Jun. to Dec. 2023 \\
Test Set     & 5000          & 4520 (90.4\%) & 480 (9.6\%)   & Jan. to Jun. 2024 
\end{tblr}
\end{table}

Each order is labeled by the issuing bank as either legitimate or fraudulent. The raw dataset exhibits significant class imbalance, with legitimate orders vastly outnumbering fraudulent ones—a characteristic typical of real-world fraud detection scenarios. To prevent data leakage from temporal overlap, we split the dataset chronologically. The training set comprises transactions from June 2023 to December 2023, while the test set covers January 2024 to June 2024. Table \ref{tab:data} provides an overview of both sets.

We adopt different balancing strategies for training and testing. For the training set, we include as many fraudulent orders as possible and mix them with a slightly larger number of legitimate orders to create an approximately balanced dataset. This balanced approach is crucial because our reinforcement learning framework uses an accuracy-based reward: with a heavily imbalanced training set, the model could exploit the reward function by trivially classifying all orders as legitimate. In contrast, the test set maintains a class distribution close to the real-world ratio. This design choice reflects the fundamental purpose of evaluation—to extrapolate how the model will perform when deployed in production, where the natural imbalance between legitimate and fraudulent transactions must be handled effectively.

\subsection{Experimental Setup}\label{chap:setup}

\textbf{Model Selection} \quad We evaluate a diverse lineup of LLMs to comprehensively assess fraud detection performance. For reinforcement learning experiments, we selected three models from the Qwen3 series \cite{yang2025qwen3}: Qwen3-4B-Instruct, Qwen3-8B, and Qwen3-14B. By choosing models of varying sizes from the same architecture family, we isolate the effect of parameter count on performance while controlling for structural differences. To establish broader performance benchmarks, we evaluate nine additional representative LLMs. Our closed-source baseline includes GPT-4.1 \cite{gpt4_1}, GPT-5-mini \cite{gpt5}, Gemini-2.5-Flash \cite{comanici2025gemini}, and Claude-4.5-Sonnet \cite{claude4_5}. For large-scale open-source models, we test Qwen3-235B-A22B-Instruct \cite{yang2025qwen3}, Kimi-K2 \cite{team2025kimi}, DeepSeek-V3 \cite{liu2024deepseek}, and GLM-4.5 \cite{zeng2025glm}. For models offering both standard and reasoning modes (e.g., extended chain-of-thought), we exclusively evaluate the standard mode, as fraud detection requires real-time responses and reasoning modes incur prohibitive latency.

\begin{figure}[t]
    \centering
    \includegraphics[width=0.98\linewidth]{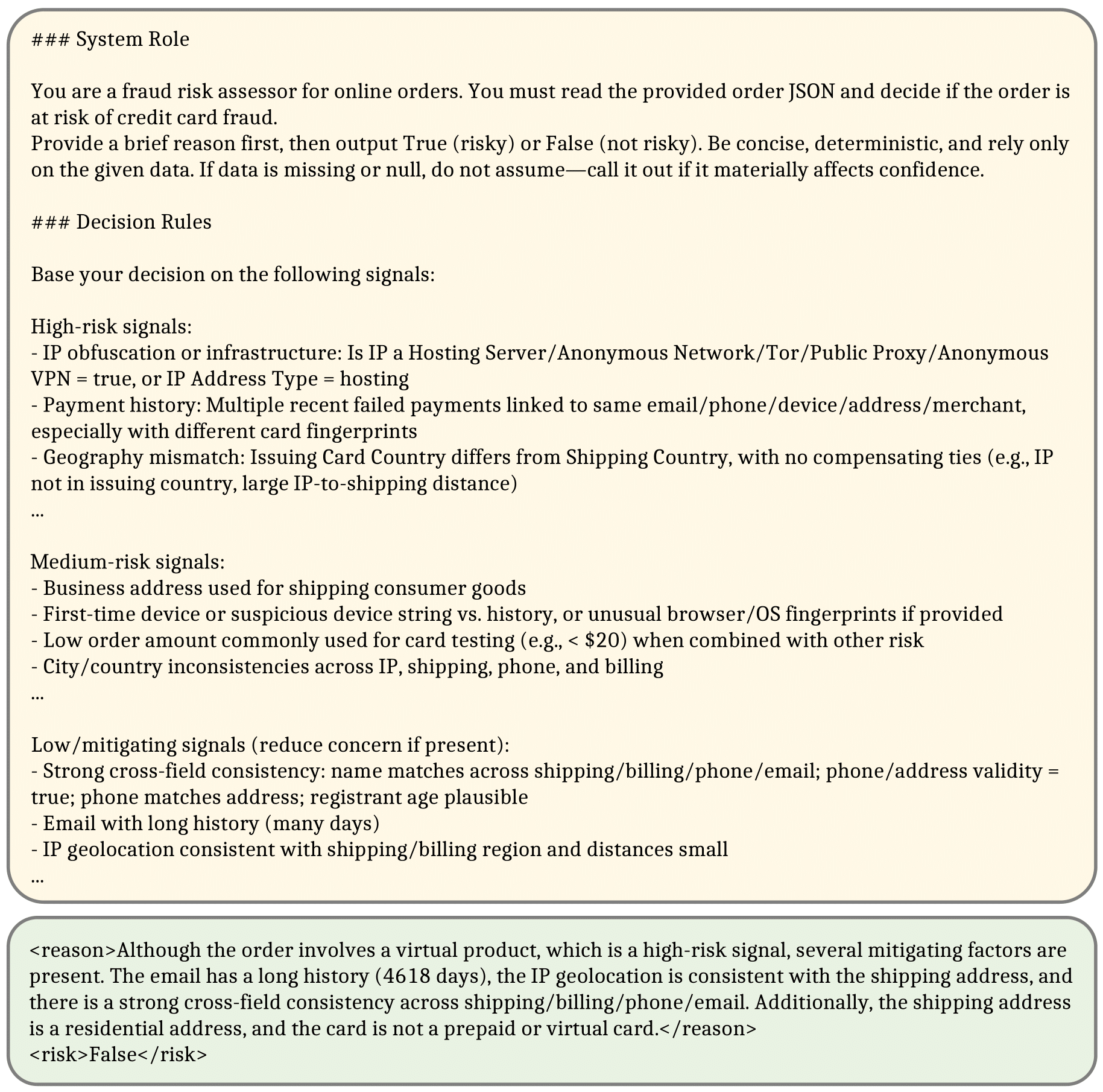}
    \caption{The prompt for the compressed setup and a response example generated by Qwen3-14B. The model is provided with a series of pre-defined risk and trust signals based on the prior experience of our anti-fraud experts, and is instructed to produce a determined, concise response. The format requirement and some defined signals are omitted in the figure for brevity.}
    \label{fig:compressed}
\end{figure}

\textbf{Training Configuration} \quad We apply the GSPO reinforcement learning algorithm to train the three Qwen3 models. Beyond the standard setup described in Section \ref{chap:RL}, we implement an alternative configuration as a control condition, which we term the \textit{compressed} version. The two configurations differ in their exploration-exploitation trade-off and response constraints:

\begin{itemize}
    \item \textbf{Standard Setup}: Uses an open-ended prompt that does not specify potential risk or trust signals, encouraging the model to explore diverse fraud indicators during training. This approach allows unconstrained exploration and reasoning in the model's responses.

    \item \textbf{Compressed Setup}: Incorporates predefined risk and trust signals directly into the prompt, leveraging prior domain knowledge to guide the model toward known fraud patterns (exploitation over exploration). Additionally, to reduce inference latency, we constrain the model to generate shorter, more deterministic responses with compressed reasoning chains.
\end{itemize}

Since this control configuration compresses both the model's exploratory behavior and reasoning verbosity, we designate it the compressed instruction setup. The prompt template and response examples for this configuration are shown in Figure \ref{fig:compressed}.

\textbf{Evaluation Metrics} \quad We employ multiple evaluation dimensions to comprehensively assess model performance in fraud detection.

\textit{Classification Performance}: We apply standard metrics for imbalanced binary classification: Accuracy, Recall (True Positive Rate), Specificity (True Negative Rate), Precision, False Positive Rate, and F1-Score. In fraud detection, misclassifications impose costs on both sides: false positives block legitimate transactions and harm merchant revenue, while false negatives allow fraudulent charges and result in bank losses. Given these symmetric costs, our primary optimization objective is maximizing F1-Score, which balances precision and recall and reflects the need for models to be neither overly conservative nor overly aggressive.

\textit{Response Efficiency}: Since credit card fraud detection requires real-time responses, we measure the average token count per completion generated by each LLM on the test set. While interpretability benefits from detailed reasoning, we prioritize reasonably concise responses that maintain clarity without excessive verbosity, as longer responses increase latency and may reduce practical deployability.

\textit{Factual Faithfulness}: A key advantage of LLMs over conventional machine learning models is their ability to produce human-interpretable explanations. However, this interpretability is only valuable if the model's reasoning faithfully reflects the input data. To assess faithfulness, we conduct hallucination testing using the DeepEval framework \cite{deepeval} on the three fine-tuned Qwen3 models. We employ Qwen3-235B-A22B-Instruct as the evaluator—a capable model with sufficient reasoning ability to serve as a reliable judge. We adopt a strict evaluation standard: a response passes the hallucination test only if the judge determines it contains absolutely no factual errors or fabricated information. Even minor inaccuracies in citing order details (e.g., incorrect amounts, mismatched addresses, or unsupported claims about historical patterns) result in test failure. This rigorous criterion ensures that deployed models provide trustworthy explanations that human reviewers can confidently rely upon.

\subsection{Results}

\subsubsection{Efficacy of Reinforcement Learning: A Comparative Analysis}\label{chap:result_1}

The quantitative evaluation of our proposed framework against baseline and state-of-the-art general-purpose models is summarized in Table \ref{tab:result}. The empirical results provide compelling evidence for the efficacy of domain-specific reinforcement learning, revealing two distinct phenomena: a radical realignment of decision boundaries in post-trained models and a significant "calibration gap" in general-purpose LLMs.

\definecolor{Gallery}{rgb}{0.941,0.941,0.941}
\begin{table}
\centering
\caption{\textbf{Performance of different LLMs on the test set.} The seven test metrics are Accuracy, Recall (True Positive Rate), Specificity (True Negative Rate), Precision, False Positive Rate, F1-Score, and average number of tokens per completion. The GSPO.C and GSPO.S denote RL with the compressed and standard instruction setup. The best F1-Score is bolded, and the second best is underlined.}
\label{tab:result}
\begin{tblr}{
  width = \linewidth,
  colspec = {Q[154]Q[148]Q[87]Q[87]Q[87]Q[87]Q[87]Q[87]Q[104]},
  row{3} = {c},
  row{8} = {c},
  row{13} = {c},
  row{14} = {Gallery},
  row{17} = {Gallery},
  row{20} = {Gallery},
  cell{1}{1} = {r=2}{},
  cell{1}{2} = {r=2}{},
  cell{1}{3} = {c=7}{0.625\linewidth,c},
  cell{3}{1} = {c=9}{0.927\linewidth},
  cell{4}{3} = {r},
  cell{4}{4} = {r},
  cell{4}{5} = {r},
  cell{4}{6} = {r},
  cell{4}{7} = {r},
  cell{4}{8} = {r},
  cell{4}{9} = {r},
  cell{5}{3} = {r},
  cell{5}{4} = {r},
  cell{5}{5} = {r},
  cell{5}{6} = {r},
  cell{5}{7} = {r},
  cell{5}{8} = {r},
  cell{5}{9} = {r},
  cell{6}{3} = {r},
  cell{6}{4} = {r},
  cell{6}{5} = {r},
  cell{6}{6} = {r},
  cell{6}{7} = {r},
  cell{6}{8} = {r},
  cell{6}{9} = {r},
  cell{7}{3} = {r},
  cell{7}{4} = {r},
  cell{7}{5} = {r},
  cell{7}{6} = {r},
  cell{7}{7} = {r},
  cell{7}{8} = {r},
  cell{7}{9} = {r},
  cell{8}{1} = {c=9}{0.927\linewidth},
  cell{9}{3} = {r},
  cell{9}{4} = {r},
  cell{9}{5} = {r},
  cell{9}{6} = {r},
  cell{9}{7} = {r},
  cell{9}{8} = {r},
  cell{9}{9} = {r},
  cell{10}{3} = {r},
  cell{10}{4} = {r},
  cell{10}{5} = {r},
  cell{10}{6} = {r},
  cell{10}{7} = {r},
  cell{10}{8} = {r},
  cell{10}{9} = {r},
  cell{11}{3} = {r},
  cell{11}{4} = {r},
  cell{11}{5} = {r},
  cell{11}{6} = {r},
  cell{11}{7} = {r},
  cell{11}{8} = {r},
  cell{11}{9} = {r},
  cell{12}{3} = {r},
  cell{12}{4} = {r},
  cell{12}{5} = {r},
  cell{12}{6} = {r},
  cell{12}{7} = {r},
  cell{12}{8} = {r},
  cell{12}{9} = {r},
  cell{13}{1} = {c=9}{0.927\linewidth},
  cell{14}{3} = {r},
  cell{14}{4} = {r},
  cell{14}{5} = {r},
  cell{14}{6} = {r},
  cell{14}{7} = {r},
  cell{14}{8} = {r},
  cell{14}{9} = {r},
  cell{15}{3} = {r},
  cell{15}{4} = {r},
  cell{15}{5} = {r},
  cell{15}{6} = {r},
  cell{15}{7} = {r},
  cell{15}{8} = {r},
  cell{15}{9} = {r},
  cell{16}{3} = {r},
  cell{16}{4} = {r},
  cell{16}{5} = {r},
  cell{16}{6} = {r},
  cell{16}{7} = {r},
  cell{16}{8} = {r},
  cell{16}{9} = {r},
  cell{17}{3} = {r},
  cell{17}{4} = {r},
  cell{17}{5} = {r},
  cell{17}{6} = {r},
  cell{17}{7} = {r},
  cell{17}{8} = {r},
  cell{17}{9} = {r},
  cell{18}{3} = {r},
  cell{18}{4} = {r},
  cell{18}{5} = {r},
  cell{18}{6} = {r},
  cell{18}{7} = {r},
  cell{18}{8} = {r},
  cell{18}{9} = {r},
  cell{19}{3} = {r},
  cell{19}{4} = {r},
  cell{19}{5} = {r},
  cell{19}{6} = {r},
  cell{19}{7} = {r},
  cell{19}{8} = {r},
  cell{19}{9} = {r},
  cell{20}{3} = {r},
  cell{20}{4} = {r},
  cell{20}{5} = {r},
  cell{20}{6} = {r},
  cell{20}{7} = {r},
  cell{20}{8} = {r},
  cell{20}{9} = {r},
  cell{21}{3} = {r},
  cell{21}{4} = {r},
  cell{21}{5} = {r},
  cell{21}{6} = {r},
  cell{21}{7} = {r},
  cell{21}{8} = {r},
  cell{21}{9} = {r},
  cell{22}{3} = {r},
  cell{22}{4} = {r},
  cell{22}{5} = {r},
  cell{22}{6} = {r},
  cell{22}{7} = {r},
  cell{22}{8} = {r},
  cell{22}{9} = {r},
  hline{1,23} = {-}{0.08em},
  hline{2} = {3-9}{0.03em},
  hline{3,8,13} = {-}{0.05em},
}
Family                                & Version      & Test Metrics &        &        &        &        &                 &          \\
                                      &              & Acc.         & TPR    & TNR    & Prec.  & FPR    & F1              & \#Tokens \\
\textit{Closed-Source LLMs}           &              &              &        &        &        &        &                 &          \\
GPT                                   & 4.1          & 0.6826       & 0.2896 & 0.7243 & 0.1004 & 0.2757 & 0.1491          & 548.8    \\
GPT                                   & 5-mini       & 0.7476       & 0.2250 & 0.8031 & 0.1082 & 0.1969 & 0.1461          & 664.9    \\
Claude                                & sonnet-4.5   & 0.6654       & 0.3083 & 0.7033 & 0.0994 & 0.2967 & 0.1503          & 725.7    \\
Gemini                                & 2.5-Flash    & 0.5244       & 0.4771 & 0.5294 & 0.0972 & 0.4706 & 0.1615          & 817.0    \\
\textit{Open-Source LLMs (\textgreater{}14B) }      &              &              &        &        &        &        &                 &          \\
Qwen3-235B                            & Instruct     & 0.4756       & 0.6021 & 0.4622 & 0.1062 & 0.5378 & 0.1806          & 668.2    \\
DeepSeek                              & v3 (671B)    & 0.6314       & 0.3554 & 0.6628 & 0.0955 & 0.3372 & 0.1487          & 449.0    \\
Kimi                                  & K2 (1T)      & 0.6880       & 0.2354 & 0.7361 & 0.0865 & 0.2639 & 0.1265          & 440.7    \\
GLM                                   & 4.5 (355B)   & 0.7090       & 0.2104 & 0.7619 & 0.0858 & 0.2381 & 0.1219          & 447.7    \\
\textit{Open-Source LLMs (4B - 14B) } &              &              &        &        &        &        &                 &          \\
Qwen3-4B                              & Instruct     & 0.6329       & 0.3708 & 0.6608 & 0.1042 & 0.3392 & 0.1627          & 761.8    \\
                                      & GSPO.C       & 0.6918       & 0.3521 & 0.7279 & 0.1208 & 0.2721 & 0.1799          & 82.3     \\
                                      & GSPO.S       & 0.8521       & 0.4313 & 0.8939 & 0.3080 & 0.1061 & \uline{0.3594}  & 282.5    \\
Qwen3-8B                              & Non-thinking & 0.2677       & 0.8438 & 0.2064 & 0.1016 & 0.7936 & 0.1814          & 626.3    \\
                                      & GSPO.C       & 0.7304       & 0.3333 & 0.7726 & 0.1347 & 0.2274 & 0.1918          & 96.3     \\
                                      & GSPO.S       & 0.8545       & 0.4250 & 0.9002 & 0.3119 & 0.0998 & \textbf{0.3598} & 447.2    \\
Qwen3-14B                             & Non-thinking & 0.4993       & 0.5292 & 0.4961 & 0.1005 & 0.5039 & 0.1689          & 522.8    \\
                                      & GSPO.C       & 0.7228       & 0.3271 & 0.7648 & 0.1287 & 0.2352 & 0.1847          & 60.7     \\
                                      & GSPO.S       & 0.8331       & 0.4604 & 0.8728 & 0.2780 & 0.1272 & 0.3467          & 370.5    
\end{tblr}
\end{table}

\begin{figure}[t]
    \centering
     \begin{subfigure}[b]{0.45\textwidth}
         \centering
         \includegraphics[width=\textwidth]{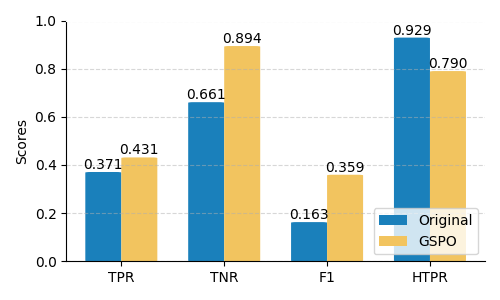}
         \caption{Qwen3-4B}
     \end{subfigure}
     \begin{subfigure}[b]{0.45\textwidth}
         \centering
         \includegraphics[width=\textwidth]{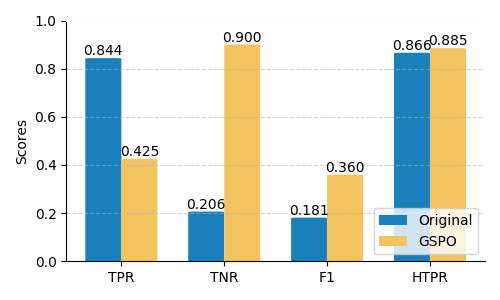}
         \caption{Qwen3-8B}
     \end{subfigure}
     \begin{subfigure}[b]{0.45\textwidth}
         \centering
         \includegraphics[width=\textwidth]{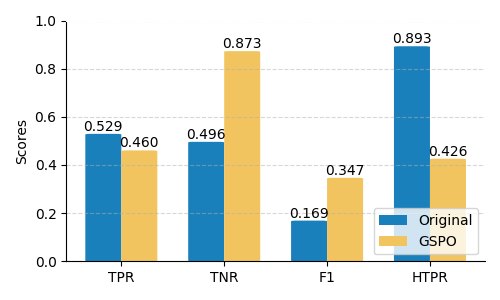}
         \caption{Qwen3-14B}
     \end{subfigure}
    \caption{Performance comparison between the original LLMs and the trained versions with GSPO and standard prompt setup. The four metrics plotted are Recall (True Positive Rate), Specificity (True Negative Rate), F1-Score, and hallucination test pass rate.}
    \label{fig:bar_chart}
\end{figure}

\textbf{The Transformative Impact of GSPO} \quad The application of GSPO reinforcement learning under the standard instruction setup (GSPO.S) induces a fundamental shift in fraud detection capabilities. As illustrated in Figure \ref{fig:bar_chart}, the post-trained Qwen3 variants achieve massive performance leaps, with the 4B, 8B, and 14B models recording F1-Score improvements of 120.90\%, 98.35\%, and 105.27\%, respectively, over their base counterparts.

Deconstructing these gains reveals that the RL process primarily optimizes performance by suppressing false positives. All three fine-tuned models exhibit a dramatic surge in Specificity (True Negative Rate)—most notably the 8B model, which improved by 336.14\%. This suggests that the untrained base models suffer from "hallucinated suspicion," frequently flagging legitimate anomalies as fraud. The reinforcement learning process effectively calibrates these models to adopt a more rigorous evidentiary standard. Interestingly, this calibration manifests differently across model sizes. The larger models (8B and 14B) adopt a conservative, precision-first posture, trading a portion of Recall (-60.50\% and -13.00\%) for maximizing Specificity. In contrast, the 4B model achieves a rare "balanced optimization," simultaneously improving both Recall (+16.32\%) and Specificity, suggesting that for this specific task complexity, the smaller parameter space may have allowed for more efficient policy exploration without overfitting to the majority class.

\textbf{The Calibration Gap in General-Purpose LLMs} \quad A striking finding from our benchmark is that our specialized, small-scale models (4B–14B parameters) consistently outperform massive state-of-the-art systems, such as Claude-4.5-Sonnet. Despite the superior reasoning capacity and world knowledge possessed by these frontier models, they fail to compete in this specific domain.

We attribute this disparity to a \textit{calibration gap} rather than a reasoning deficiency. General-purpose LLMs can oftentimes successfully identify relevant signals—noting that an IP address is a proxy or that a shipping distance is large—but they lack the domain-specific experience to weigh these signals against a latent decision threshold. Figure \ref{fig:learning} visualizes the critical mechanism that bridges this gap: the reinforcement learning loop. Through the backpropagation of reward signals derived from ground truth, our models implicitly attune their latent decision boundaries to the specific distribution of the financial dataset. Without this targeted optimization, a general LLM cannot determine if a "500km distance" is a high-risk anomaly or a standard deviation for this specific merchant category. By fine-tuning via GSPO, we effectively anchor the reasoning capabilities of the smaller models to the specific risk landscape of the financial dataset, allowing them to surpass models that are orders of magnitude larger but uncalibrated to the task.

\textbf{Emergent Efficiency and Information Density} \quad A critical, emergent property of our reinforcement learning pipeline is the spontaneous optimization of response length. Without explicit penalization for verbosity, the GSPO process drove the 4B, 8B, and 14B models to compress their outputs by 62.92\%, 28.60\%, and 29.13\%, respectively. This reduction does not represent a loss of information, but rather an increase in \textit{information density}: the models learned to discard superfluous linguistic filler and focus exclusively on high-leverage diagnostic signals. This "information distillation" is operationally vital; by naturally converging on concise reasoning paths, the models significantly reduce inference latency, directly facilitating deployment in high-throughput, real-time financial transaction systems where millisecond-level delays are prohibitive.

\begin{figure}[t]
    \centering
    \includegraphics[width=0.98\linewidth]{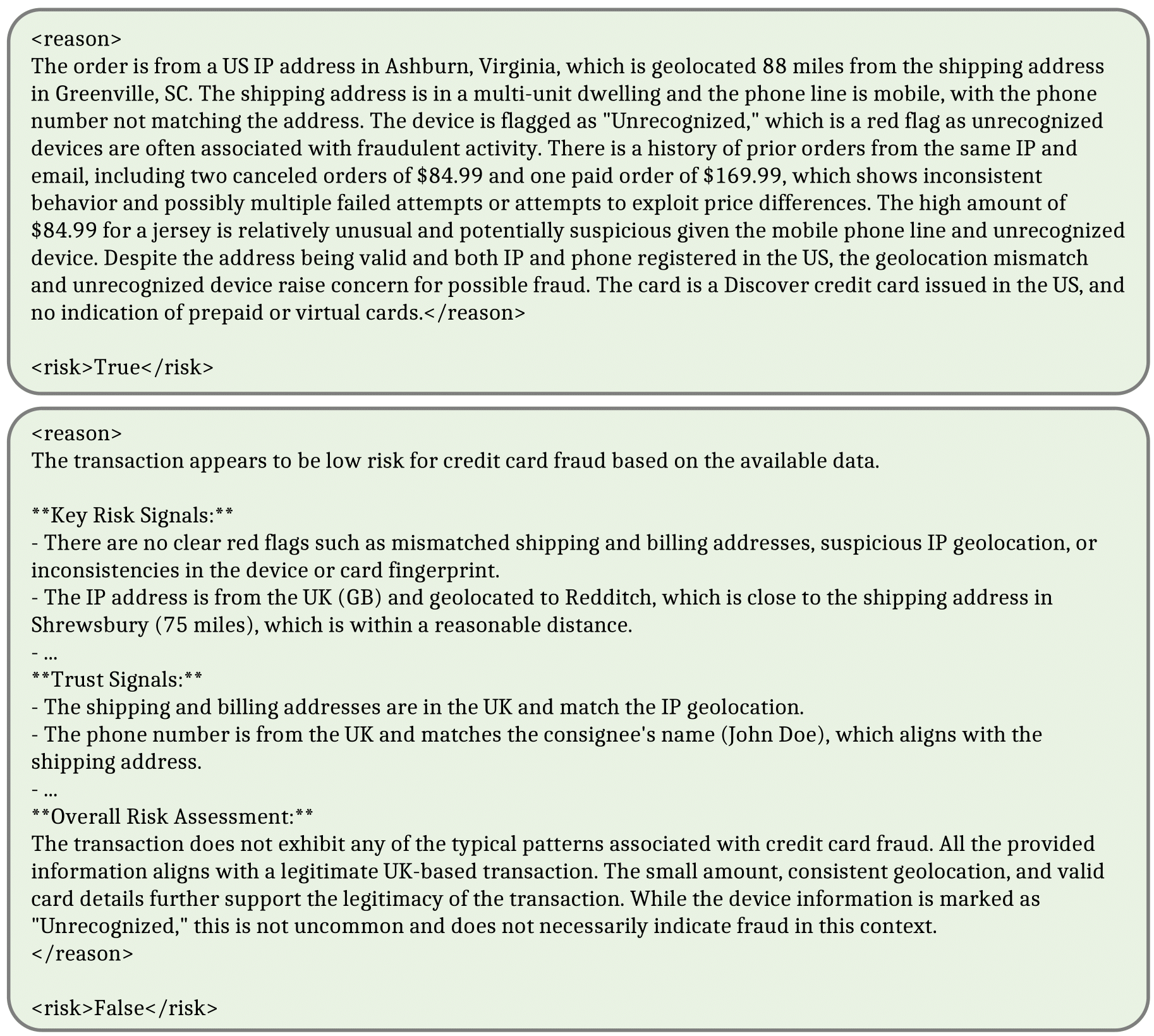}
    \caption{Example outputs of the trained Qwen3-4B and Qwen-8B, respectively. Some signals identified are omitted for brevity. Models of different scales tend to converge to different reasoning styles despite the same instruction. }
    \label{fig:example_output_more}
\end{figure}

\textbf{Divergence in Cognitive Strategies} \quad Despite sharing the same architecture, prompt template, and optimization objective, the three models converged onto distinct "cognitive strategies" for fraud analysis (see Figure \ref{fig:example} and \ref{fig:example_output_more}). \begin{itemize} 
    \item \textbf{The Deductive Strategist (14B):} The Qwen3-14B adheres strictly to a logical flow, systematically isolating risk and trust signals before synthesizing them into a verdict. This mirrors standard Chain-of-Thought (CoT) reasoning. 
    \item \textbf{The Executive Strategist (8B):} The Qwen3-8B adopts a "verdict-first" structure, declaring the classification immediately and providing retrospective justification. This suggests the model has internalized the decision boundary so effectively that it requires less "thinking time" to reach a conclusion. 
    \item \textbf{The Holistic Strategist (4B):} The Qwen3-4B integrates evidence in a free-form narrative, blending signal extraction and judgment into a cohesive paragraph. 
\end{itemize} 

This policy divergence indicates that the solution space for fraud detection contains multiple
local optima; there is no single "correct" way to reason about fraud, and models of different capacities navigate this landscape differently to maximize reward.

\textbf{Inverse Scaling of Faithfulness: The Capacity Trap} \quad Perhaps the most counter-intuitive finding is the inverse relationship between model scale and factual faithfulness under RL finetuning. Contrary to standard scaling laws, larger models did not yield superior holistic performance. The 4B and 8B models maintain a comparable hallucination test pass rate before and after training (over 80\%). In contrast, while the 14B model improved its F1-Score, it suffered a catastrophic 46.7 percentage point drop in the hallucination pass rate (Figure \ref{fig:bar_chart}), vastly underperforming the 4B and 8B models in truthfulness.

We interpret this as a manifestation of "reward hacking" facilitated by higher model capacity. The larger 14B model, possessing superior linguistic fluency and world knowledge, appears to utilize these capabilities to \textit{fabricate} convincing-sounding evidence to justify its predictions, rather than relying solely on the input data. Essentially, the model prioritizes \textit{persuasiveness} (to maximize the reward) over \textit{truthfulness}. In contrast, the smaller models, constrained by lower capacity, are less adept at hallucinating plausible details and thus remain more grounded in the provided context. This "Less Is More" phenomenon highlights a critical safety challenge: in high-stakes financial domains, increased model intelligence can paradoxically lead to less reliable explanations if the reward signal does not explicitly penalize fabrication.

\begin{figure}[t]
    \centering
    \includegraphics[width=0.98\linewidth]{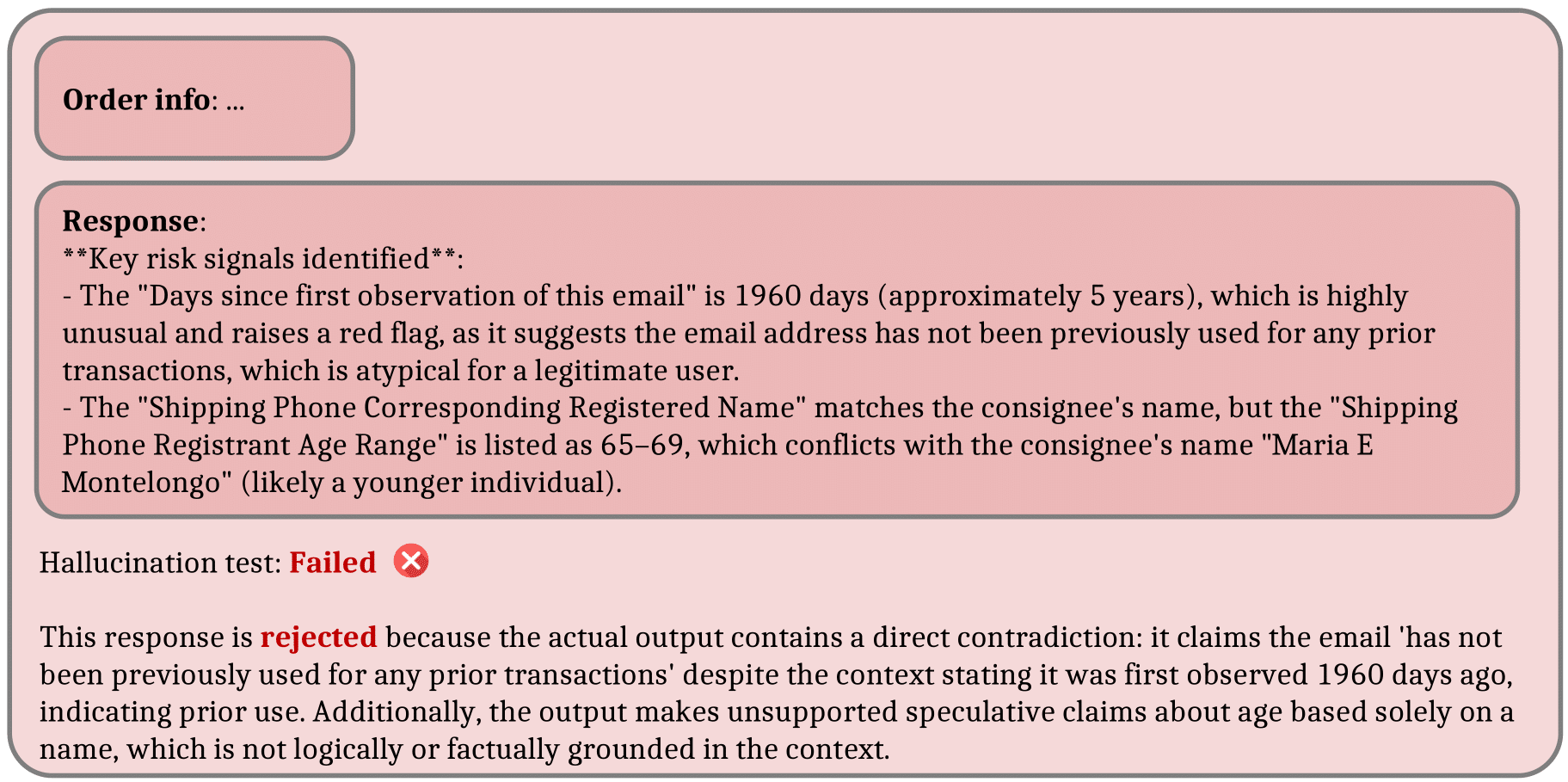}
    \caption{A response generated by the trained Qwen3-14B that fails the hallucination test, due to factual errors identified by the judge.}
    \label{fig:hallucination}
\end{figure}

\subsubsection{The Critical Role of Exploration and Cognitive Space}\label{chap:result_2}

\begin{figure}[t]
    \centering
     \begin{subfigure}[b]{0.47\textwidth}
         \centering
         \includegraphics[width=\textwidth]{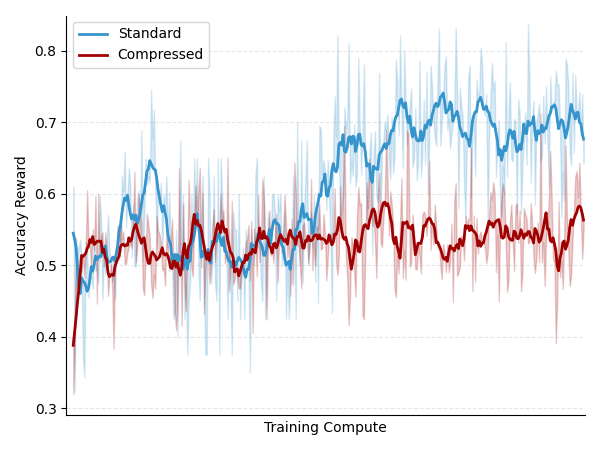}
         \caption{Accuracy Reward}
     \end{subfigure}
    \begin{subfigure}[b]{0.47\textwidth}
         \centering
         \includegraphics[width=\textwidth]{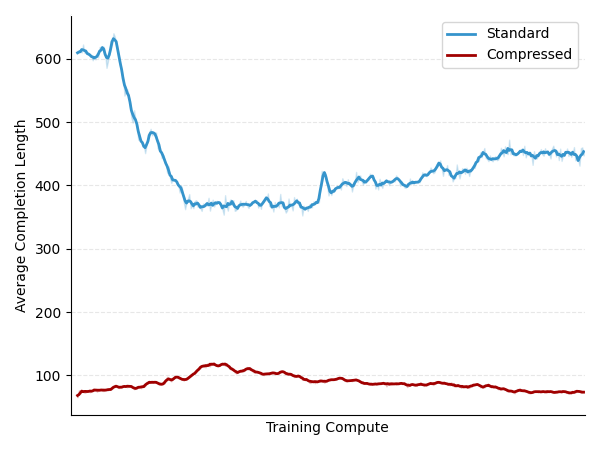}
         \caption{Average Completion Length}
     \end{subfigure}
    \caption{Comparison between the standard and compressed instruction setup on the accuracy reward and average completion length during Qwen3-8B GSPO training. The x-axis is training compute, which is equivalent to the number of samples processed in RL. The plot is smoothed with moving averages, and the shaded area represents the actual value of the plotted metric.}
    \label{fig:curve}
\end{figure}

\textbf{The Cost of Premature Constraints} \quad To isolate the impact of reasoning freedom during the reinforcement learning process, we conducted a comparative ablation between the open-ended "Standard" setup and the "Compressed" variant (described in Section \ref{chap:setup}). The Compressed setup represents a "human-guided" approach: it injects domain knowledge (predefined signals) and strictly enforces brevity to minimize latency.

The results, presented in Table \ref{tab:result} and Figure \ref{fig:curve}, reveal a stark trade-off. While the Compressed configuration successfully minimized token usage, it caused a functional collapse in learning efficacy. The 4B, 8B, and 14B models achieved trivial F1-Score gains of 10.57\%, 5.73\%, and 9.35\%, respectively. In sharp contrast, the Standard setup—which permitted unconstrained exploration—delivered performance gains over an order of magnitude higher (ranging from 98.35\% to 120.90\%).

\textbf{The "Cognitive Straitjacket" Effect} \quad We attribute this massive performance gap to two mechanisms: the suppression of exploration and the truncation of reasoning computation. First, by hard-coding specific risk signals into the prompt, the Compressed setup imposes a "cognitive straitjacket" on the model. It biases the optimization process toward exploiting known heuristics rather than discovering novel, data-driven patterns. The Standard setup, conversely, allows the model to treat the feature space as a \textit{tabula rasa}, enabling the RL agent to identify subtle, non-intuitive correlations that human designers might overlook.

Second, the failure of the Compressed models provides strong empirical support for the "Chain-of-Thought" hypothesis in the context of fraud detection. Large Language Models perform computation via autoregressive token generation; essentially, "thinking" requires "speaking." By forcing the Compressed models to output a verdict immediately or with minimal elaboration, we effectively slashed their computational budget for reasoning. The superior performance of the Standard setup confirms that accurate fraud classification is not merely pattern matching—it requires a sequence of intermediate reasoning steps to weigh conflicting evidence. Prematurely truncating this chain forces the model to guess rather than deduce.

\textbf{Implications for Real-Time Systems} \quad These findings resolve a fundamental tension in deploying LLMs for financial risk control. While low latency is non-negotiable, our results demonstrate that \textit{enforced} brevity is destructive. Instead, the optimal strategy is \textit{organic} efficiency: using the Standard setup allows the model to self-optimize, naturally discarding useless tokens to maximize reward (as observed in Section \ref{chap:result_1}) while retaining the necessary "cognitive space" to reason through complex fraud scenarios.

\section{Discussion}\label{chap:discussion}

As mentioned in Section \ref{chap:result_1}, we discover a "less is more" phenomenon in our tailored fraud detection task, where smaller LMs may outperform larger LMs after the same RL training. We hypothesize several contributing factors:

\textit{Capacity-Task Alignment}: For highly specialized tasks like fraud classification, smaller models' limited capacity may paradoxically be advantageous. With fewer parameters, these models can be more effectively molded to dataset-specific patterns without interference from extensive pre-existing knowledge that may be irrelevant or even counterproductive for the narrow task at hand.

\textit{Catastrophic Forgetting}: Larger models possess more extensive and complex learned representations from pretraining. During RL fine-tuning on a narrow task, the intense optimization pressure may cause these models to overwrite or discard general reasoning capabilities that, while not directly applicable to fraud detection, provided a stable foundation for factual grounding \cite{li2024revisiting}. This manifests as increased hallucination—the model becomes overfitted to fraud patterns while degrading its ability to faithfully represent input information.

\textit{Optimization Dynamics}: Smaller models may exhibit more stable gradient landscapes during RL training on limited data, allowing them to converge to better local optima that balance task performance with faithfulness to input data.

These findings suggest that for production deployment of LLM-based fraud detection systems, practitioners should consider the full Pareto frontier of model size, task performance, response efficiency, and factual reliability, rather than defaulting to the largest available model.

While our study focuses exclusively on LLM-based classification, the complementary strengths of language models and conventional machine learning suggest promising opportunities for hybrid systems. Traditional ML models excel at processing structured numerical features and capturing statistical patterns across large-scale tabular data with minimal latency. In contrast, LLMs demonstrate superior ability to reason about textual information, behavioral patterns, identity inconsistencies, and complex multi-modal signals that resist traditional feature engineering. Future work should explore optimal integration strategies, including cascaded architectures, ensemble methods, and dynamic routing mechanisms that allocate cases based on complexity. Also, we observe degradation in factual faithfulness after RL training, especially for larger LLMs. This poses risks for production deployment, as fraud analysts must trust model explanations when reviewing flagged transactions. Future work can consider mitigating this issue by incorporating explicit faithfulness rewards alongside accuracy rewards during RL training, potentially using automated fact-checking systems to penalize fabricated details.

\section{Related Works}

\textbf{Large Language Models in Financial Fraud Detection} \quad Traditionally, credit card fraud detection has been dominated by supervised machine learning models such as Support Vector Machine\cite{gyamfi2018bank, li2021comparative, singh2022financial}, Random Forests\cite{xuan2018random, jonnalagadda2019credit}, Gradient Boosting Decision Trees (XGBoost \cite{hajek2023fraud, noviandy2023credit, baisholan2025fraudx} and LightGBM \cite{zhao2024improved, xiao2025application}), and Neural Networks \cite{alghofaili2020financial, chen2021deep, udayakumar2023deep}, which excel at processing structured tabular data. While highly accurate, these "black-box" models often lack semantic understanding and interpretability, a critical requirement in financial compliance, calling for a line of studies in interpretable fraud detection \cite{rudin2019stop, lin2022model, hasan2024explainable, li2024sefraud, baisholan2025fraudx}. Recent research has begun to explore LLMs for this domain, leveraging their natural language processing capability. For example, Ren et al. \cite{ren2025ai} investigated LLM-based fraud detection with a multi-agent system in complex, realistic scenarios. Singh et al. \cite{singh2025advanced} have employed Retrieval-Augmented Generation (RAG) to provide LLMs with relevant context for real-time fraud detection. Huang et al. \cite{huang2025can} and Yang et al. \cite{yang2025flag} incorporate LLMs in Graph Neural Networks for semantic analysis in fraud detection. However, these methods typically treat the LLM as a frozen inference engine, limiting its ability to internalize the complex, high-dimensional patterns of fraudulent behavior found in massive transaction datasets.

\textbf{Reinforcement Learning for LLM Post-Training} \quad Reinforcement Learning from Human Feedback (RLHF) has become the standard for aligning LLMs with complex objectives \cite{wang2024comprehensive}. Standard algorithms like Proximal Policy Optimization (PPO) \cite{schulman2017proximal} and Direct Preference Optimization (DPO) \cite{rafailov2023direct} have been widely used to improve instruction following. However, standard PPO often suffers from training instability and high memory costs, because it trains a policy and a critic model at the same time, which are typically of the same scale. To address this, the Group Relative Policy Optimization (GRPO) algorithm \cite{shao2024deepseekmath} was introduced, eliminating the need for a critic model by normalizing rewards within a group of outputs. The more recent Group Sequence Policy Optimization (GSPO) \cite{zheng2025group} further stabilizes training by optimizing at the sequence level rather than the token level. RL represents a shift from general instruction tuning to task-specific behavioral alignment, allowing the model to "learn" the subtle risk signals in transaction data through trial-and-error rather than just supervised imitation.

\section{Conclusion}

This work proposes a method that leverages the textual understanding and reasoning capabilities of Large Language Models combined with Reinforcement Learning for credit card fraud detection using only raw transaction data. Compared to conventional machine learning approaches, this method provides intuitive, natural language interpretability. Our results demonstrate that \textbf{fine-tuning lightweight language models with RL yields substantial performance gains in fraud detection tasks while preserving interpretability}. Across multiple Qwen3 models, task-specific RL with thoughtfully designed rewards improves F1-Score and related metrics, often with shorter outputs suitable for low-latency deployment. Importantly, we observe a “less is more” pattern: smaller, task-adapted models can match or exceed the performance of larger counterparts after RL, highlighting the value of domain-focused optimization. We also highlight the need to balance exploration and faithfulness. Open-ended signal discovery enhances robustness and explanatory richness, but excessive compression or over-reliance on pre-defined signals can extensively degrade discrimination. Practical deployment thus favors a hybrid approach: continue refining faithfulness-aware objectives and verification, while exploring hybrid architectures that combine LLMs’ interpretability with traditional models’ speed for structured data.

\bibliographystyle{unsrt}  
\bibliography{references}  






\end{document}